\documentclass[runningheads]{llncs}
\usepackage[T1]{fontenc}
\usepackage{graphicx,verbatim}
\usepackage{xcolor}
\usepackage{graphics}
\usepackage{epsfig} 
\usepackage{mathptmx} 
\usepackage{times}
\usepackage{amsmath,amssymb,amsfonts,bm}
\usepackage{algorithm,algorithmic}
\usepackage{flushend}
\usepackage{booktabs,threeparttable,multirow, caption}
\usepackage{color} 
\usepackage{cuted}
\usepackage{array}
\usepackage{tabu}                     
\usepackage{multirow}                
\usepackage{multicol}               
\usepackage{multirow}              
\usepackage{float}                   
\usepackage{makecell}               
\usepackage{booktabs} 
\usepackage{marvosym}
\usepackage{hyperref}

\usepackage{color}

\begin{document}
\title{Harnessing Foundation Models for Robust and Generalizable 6-DOF Bronchoscopy Localization}
\titlerunning{Harnessing Foundation Models for Bronchoscopy Localization}
%


\author{Qingyao Tian\inst{1,2}\and
Huai Liao\inst{3}\and
Xinyan Huang\inst{3}\and
Bingyu Yang\inst{1,2}\and
Hongbin Liu\inst{1,4,5}\textsuperscript{\Letter}
}
\authorrunning{Q. Tian et al.}

\institute{State Key Laboratory of Multimodal Artificial Intelligence Systems, Institute of Automation, Chinese Academy of Sciences, Beijing, China
\email{liuhongbin@ia.ac.cn}\and
School of Artificial Intelligence, University of Chinese Academy of Sciences, Beijing, China\and
The First Affiliated Hospital, Sun Yat-sen University, Guangzhou, China\and
Centre for Artificial Intelligence and Robotics, Chinese Academy of Sciences, HK, China\and
School of Engineering and Imaging Sciences, King’s College London, UK}
\maketitle              
\begin{abstract}
Vision-based 6-DOF bronchoscopy localization offers a promising solution for accurate and cost-effective interventional guidance. However, existing methods struggle with 1) \textit{limited generalization} across patient cases due to scarce labeled data, and 2) \textit{poor robustness} under visual degradation, as bronchoscopy procedures frequently involve artifacts such as occlusions and motion blur that impair visual information. To address these challenges, we propose \textbf{PANSv2}, a generalizable and robust bronchoscopy localization framework. Motivated by PANS \cite{tian2024pans} that leverages multiple visual cues for pose likelihood measurement, PANSv2 integrates depth estimation, landmark detection, and centerline constraints  into a unified pose optimization framework that evaluates pose probability and solves for the optimal bronchoscope pose. To further enhance generalization capabilities, we leverage the endoscopic foundation model EndoOmni~\cite{tian2024endoomni} for depth estimation and the video foundation model EndoMamba~\cite{tian2025endomamba} for landmark detection, incorporating both spatial and temporal analyses. Pretrained on diverse endoscopic datasets, these models provide stable and transferable visual representations, enabling reliable performance across varied bronchoscopy scenarios. Additionally, to improve robustness to visual degradation, we introduce an automatic re-initialization module that detects tracking failures and re-establishes pose using landmark detections once clear views are available. Experimental results on bronchoscopy dataset encompassing 10 patient cases show that PANSv2 achieves the highest tracking success rate, with an 18.1\% improvement in SR-5 (percentage of absolute trajectory error under 5 mm) compared to existing methods, showing potential towards real clinical usage.

\keywords{Surgical navigation  \and Foundation model \and Bronchoscopy localization}
\end{abstract}
\section{Introduction}
Bronchoscopy is widely used for visual inspection, diagnosis, and biopsy of pulmonary lesions \cite{andolfi2016role}. During the procedure, clinicians navigate a camera-integrated flexible endoscope through the bronchial tree, guided by pre-operative CT scans to reach target regions such as peripheral airways and pulmonary nodules. However, the endoscope’s limited field of view, coupled with the complex anatomy of the airway, makes accurate localization challenging and highly dependent on operator expertise. To address this, there is growing interest in developing automatic 6-DoF bronchoscope localization methods to enable robust and efficient navigation support \cite{kops2023diagnostic,cold2024artificial}.

Recent approaches for vision-based localization have explored a range of techniques, including image retrieval \cite{zhao2019generative,sganga2019autonomous}, visual odometry \cite{deng2023feature,borrego2023bronchopose}, landmark detection \cite{keuth2024airway,tian2024bronchotrack}, and hybrid strategies incorporating rule-based logic \cite{tian2024dd} or probabilistic models \cite{tian2024pans}. While these methods show promise for cost-effective, image-based localization, their deployment in real-world clinical workflows is limited by two key challenges: (1) \textit{poor generalization across patients}, and (2) \textit{limited robustness to visual degradation}. The lack of generalization is primarily due to insufficient training data, as collecting large-scale, accurately labeled bronchoscopic pose annotations is labor-intensive and time-consuming. In terms of robustness, bronchoscopic videos are frequently degraded by visual occlusions such as fluids, bubbles, and motion blur. These artifacts severely impact visual feature extraction, often leading to failure cases. Consequently, many prior works filter out low-quality frames during evaluation \cite{tian2024dd,tian2024bronchotrack}, or require frequent re-calibration to reinitialize tracking \cite{tian2024pans}.

In this work, we present PANSv2, a robust framework for 6-DoF bronchoscope localization. Building upon the original PANS~\cite{tian2024pans}, which fuses multiple visual cues to enhance localization robustness, PANSv2 integrates depth estimation, anatomical landmark detection, and centerline constraints to evaluate the probability of candidate poses and solve for the optimal pose that aligns the virtual bronchoscopic view with the real one. To overcome the generalization challenge, we construct the largest bronchoscopic localization dataset to date, comprising 66 procedures and over 30k labeled frames, providing a strong foundation for data-driven learning. To further enhance generalization, we incorporate foundation models pretrained on large-scale endoscopic datasets: EndoOmni \cite{tian2024endoomni} for depth estimation and EndoMamba \cite{tian2025endomamba} for landmark detection from video input. These models offer stable and transferable visual representations, enabling consistent performance across diverse patient anatomies and imaging conditions. To improve robustness under visual degradation, we introduce an automatic re-initialization module that detects tracking failures and recovers pose using landmark detections when clear views return. As a result, our method achieves significantly improved tracking continuity under intervention dataset captured under real clinical conditions. Evaluations on 10 full-length bronchoscopy procedures show that PANSv2 achieves the highest tracking success rate among existing methods, demonstrating strong potential for clinical deployment.

\begin{figure*}[tbp]
\centerline{\includegraphics[width=\textwidth]{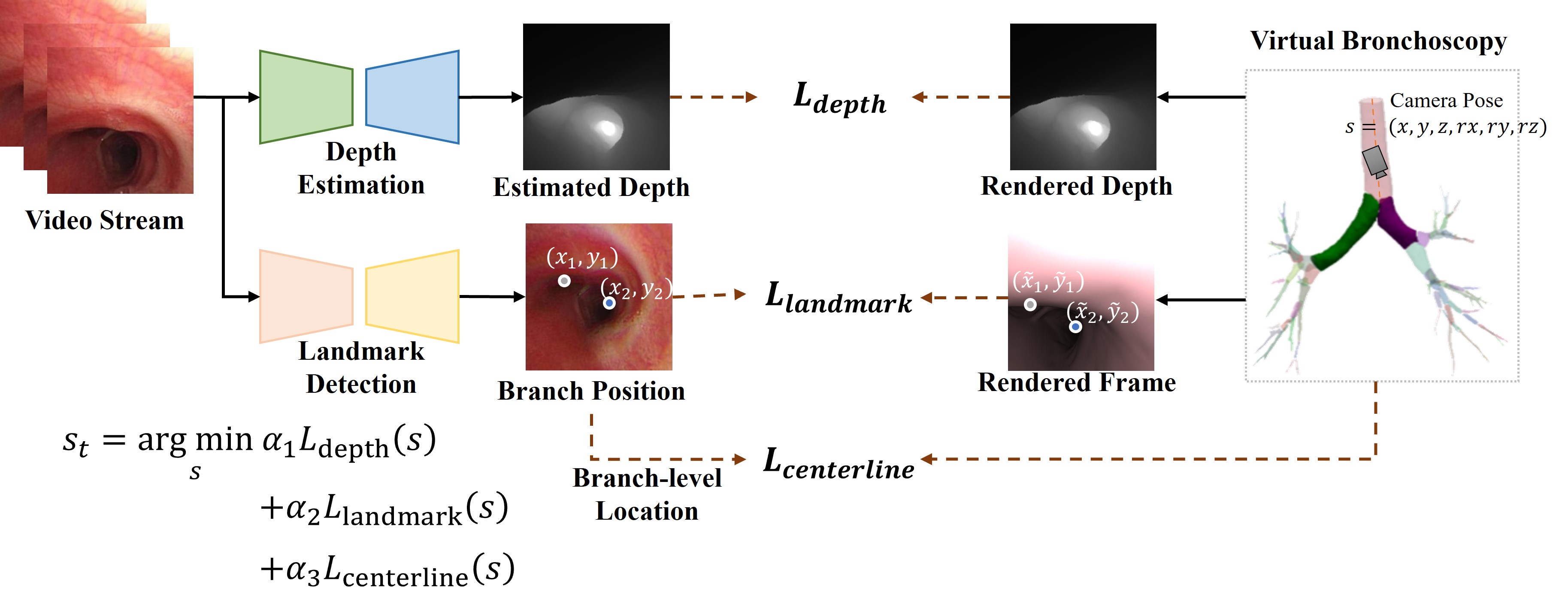}}
\caption{Overview of the proposed PANSv2 for robust bronchoscope localization. The bronchoscope pose $s_t$ at time $t$ is estimated by optimizing the depth similarity, landmark alignment and centerline constraint.}
\label{fig1}
\end{figure*}

\section{Methods}
\subsection{Overview}
Our PANSv2 framework is illustrated in Fig. \ref{fig1}. The core of bronchoscopic localization is to align the view from a virtual bronchoscopic camera—rendered within the segmented airway from CT scans—with the real bronchoscopic image, thereby estimating the real camera’s pose in the CT coordinate system. To achieve this, we formulate a probabilistic cost function that evaluates the likelihood of a given camera pose $s$. This cost $L$ is computed based on three cues: depth similarity, alignment of landmark image coordinates, and a centerline constraint. Formally, the cost is formulated as:

\begin{equation}
    L(s) = \alpha_1 L_{\mathrm{depth}}(s) + \alpha_2 L_{\mathrm{landmark}}(s) + \alpha_3 L_{\mathrm{centerline}}(s).
    \label{eq:total_loss}
\end{equation}

By solving for the camera pose that minimizes this cost, we obtain the estimated pose $s_t$ of the bronchoscope at each time $t$:

\begin{equation}
    s_t = \mathop{\arg\min}\limits_{s}{L(s)}.
    \label{eq:optimize}
\end{equation}

Given the initial bronchoscopic pose at the first time step, at each following time step, our PANSv2 solves optimized bronchoscopic camera pose by Powell \cite{fletcher1963rapidly}, using the previous estimated pose as initial value.

\subsection{Depth Similarity}
To align the virtual bronchoscopic view with the real one, we first estimate the cost of a candidate camera pose by measuring the similarity between the depth map rendered in the virtual airway and the depth estimated from the real bronchoscopic frame. By aligning depth rather than RGB images, we eliminate texture-specific appearance variations in the airway and focus solely on geometric structure, enabling more robust and interpretable alignment.

To estimate depth from bronchoscopic frames, we employ EndoOmni \cite{tian2024endoomni}, a foundation model for endoscopic depth estimation trained on over 700,000 frames from a wide range of procedures and sources. Since EndoOmni predicts relative depth without an absolute scale, we use Normalized Cross-Correlation (NCC) to quantify similarity between the estimated and rendered depths. Formally:

\begin{equation}
L_{\mathrm{depth}}(s) = 1 - \mathrm{NCC}(z, \bar{z}(s)) = 1 - \frac{ \sum_i (z_i - \mu_{z})(\bar{z}_i(s) - \mu_{\bar{z}(s)}) }{ \sqrt{ \sum_i (z_i - \mu_{z})^2 } \sqrt{ \sum_i (\bar{z}_i(s) - \mu_{\bar{z}(s)})^2 } },
\end{equation}

\noindent where $z$ denotes the estimated depth map of the real bronchoscopic frame, $\bar{z}(s)$ is the depth rendered from the virtual camera at pose $s$, and $\mu_{z}, \mu_{\bar{z}(s)}$ are the mean depths of the respective maps. The cost $L_{\mathrm{depth}}$ is minimized when the two depth maps are maximally correlated, corresponding to high geometric alignment.

\subsection{Landmark Alignment}
Different sections of the human airway often share similar geometric structures, such as tubular shapes or bifurcations. As a result, relying solely on depth similarity can introduce ambiguity when optimizing the camera pose. To address this, PANSv2 leverages EndoMamba \cite{tian2025endomamba}, an efficient endoscopic video foundation model, to detect anatomical landmarks within the bronchoscopic view.  Built on a Mamba-based backbone for joint spatial and temporal modeling, EndoMamba leverages information from both the current frame and preceding frames to enhance landmark detection. This temporal context helps disambiguate visually similar airway regions, improving the accuracy and reliability of landmark-based pose estimation. Formally, given a bronchoscopic image $I_t$ and hidden states from previous time points $h_{t-1}$, the landmark detection model outputs:

\begin{equation}
    f_{\text{landmark}}(I_t, h_{t-1}) = (\mathbf{M}_t, h_t), 
\end{equation}

\begin{equation}
    \quad \mathbf{M}_t \in \mathbb{R}^{n \times 3}, \quad 
    \mathbf{M}_{t,i} = (v_i, x_i, y_i),
\end{equation}

\noindent where $ v_i \in [0, 1] $ is the predicted visibility score, and $ (x_i, y_i) $ are the 2D image coordinates of landmark $ i $, for $ i = 1, \dots, n $, with $n$ being the number of defined anatomical landmarks.

To evaluate how well a candidate camera pose aligns with observed anatomy, we compare the detected landmark positions from the model with the expected projections of known anatomical landmarks from the CT-based virtual model. Specifically, for a given camera pose $ s $, we project the 3D landmark positions from the CT into image space, yielding the set of expected 2D coordinates $ \{ (\tilde{x}_i(s), \tilde{y}_i(s)) \}_{i=1}^n $. The landmark alignment cost is then defined as the average L2 distance between the detected and projected positions of visible landmarks:

\begin{equation}
    L_{\text{landmark}}(s) = \frac{1}{\sum_i v_i} \sum_{i=1}^n v_i \cdot \left\| (x_i, y_i) - (\bar{x}_i(s), \bar{y}_i(s)) \right\|_2,
\end{equation}

\noindent where $(x_i, y_i) $ and $v_i$ are the image coordinates and visibility predicted by the landmark detection model, and $(\bar{x}_i(s), \bar{y}_i(s)) $ is the projection of landmark $i $ given pose $s$. The loss is computed only over visible landmarks, as indicated by $v_i = 1 $.

\subsection{Centerline Constraint}
Preliminary experiments show that using only depth and landmark alignment for pose evaluation often results in camera poses drifting outside the airway mesh. To address this issue, and to constrain the search space for improved convergence, PANSv2 incorporates a centerline-based constraint as part of the pose cost function. The centerline constraint is motivated by the assumption that the bronchoscopic camera should remain close to the airway centerline and generally align its viewing direction with the local orientation of the airway. To implement this, we first infer the branch-level location of the bronchoscope based on detected landmarks with a voting strategy following \cite{tian2024bronchotrack}. Then, given branch-level location $ b = 1, \dots, n$, where $n$ being the number of defined anatomical branches, the centerline cost is defined as:

\begin{equation}
    L_{\mathrm{centerline}}(s) = \mathcal{N}(d;\, 0, \sigma_1^2) \cdot \mathcal{N}(\phi;\, 0, \sigma_2^2),
\end{equation}

\noindent where $d$ is the distance from camera pose $s$ to branch $b$, and $\phi$ represents the angle between pose $s$ and branch $b$. The variances $\sigma_1$ and $\sigma_2$ are set as $\frac{r}{2}$ and $\frac{\pi}{6}$, with $r$ being the radius of the branch $b$.

\subsection{Automatic Re-initialization}
During bronchoscopy interventions, it is common for surgeons to navigate the bronchoscope into a new airway branch, even when the image is temporarily degraded by severe visual artifacts. In such cases, pose optimization often fails during the artifact period and cannot recover afterward without external intervention, since the previous estimated pose (used as the initialization for optimization) is far from the bronchoscope's true current location.

To address this, PANSv2 adopts a simple yet effective strategy for automatic pose re-initialization. We first detect optimization failure by thresholding the final pose cost:

\begin{equation}
    \delta_{\text{fail}}(s) = 
\begin{cases}
\text{True}, & \text{if } L(s) > \tau \\
\text{False}, & \text{otherwise}
\end{cases}.
\end{equation}

After the field of view recovers, EndoMamba recognizes visible anatomical landmarks and determines the current branch-level location $b$ of the bronchoscope. We then reinitialize the pose optimization by setting the initial pose to the center point of branch $b$'s centerline segment, and continue with the optimization process with Eq. \ref{eq:optimize}.

\section{Experiments}
\subsection{Implementation Details}
\textbf{Dataset.} We constructed a bronchoscopy localization dataset consisting of 66 cases, with 56 cases for training and 10 for testing. Each case includes the patient-specific CT scan, camera checkerboard calibration, and a video clip containing approximately 1500 to 3500 frames. The camera poses for each bronchoscopic frame are annotated, except when the pose cannot be determined due to severe occlusions. Note that frames without pose annotations are still processed by PANSv2 but do not contribute to metric calculations. Localization annotations were generated using our OpenGL-based toolkit, which aligns virtual camera intrinsics with the actual bronchoscope. Three experts manually labeled the dataset by registering the virtual views to the real data. To assess annotation accuracy, two cases were labeled independently by the experts, yielding a group variance of 0.58 mm. After obtaining coordinate-level annotations, we generate branch landmark detection labels using the airway centerline. A branch is considered visible if any point along its centerline is visible from the current camera pose, with the 2D position of the branch defined as the furthest visible centerline point in the image.

\textbf{Foundation models.} For depth estimation, we use the pre-trained EndoOmni-b without fine-tuning for zero-shot scale-and-shift-invariant depth estimation. Each video frame is resized to 378×378 as input to EndoOmni. To detect visible branches in each frame, we sample 16-frame input clips of spatial size 224×224 for fine-tuning EndoMamba. The learning rate is set to 1e-6 with 20 training epochs.

\textbf{Evaluation Metrics.} Following previous work \cite{shen2019context,Gu2022}, we use Absolute Trajectory Error (ATE), SR-5 (percentage of ATE < 5 mm), and SR-10 (percentage of ATE < 10 mm) to evaluate our method. Among these, SR-5 and SR-10 serve as indicators of the success rate for endoscope tracking.

\subsection{Comparison with State of the Art}
We compare PANSv2 with existing state-of-the-art (SOTA) monocular endoscopic localization techniques. This includes depth registration (Depth-Reg) approaches \cite{shen2019context,banach2021visually}, the visual-odometry-based method Endo-Depth-and-Motion \cite{recasens2021}, the hybrid method DD-VNB \cite{tian2024dd}, and the previous version, PANS \cite{tian2024pans}. Since prior methods exclude frames affected by visual degradation—caused by severe airway deformation or rapid endoscope withdrawal—we additionally evaluate PANSv2 on these filtered frames. Notably, PANSv2 processes the complete set of unfiltered video frames using only a single initialization at the trachea, whereas the compared methods require two separate initialization steps—one before entering each side of the airway.

\begin{figure*}[tbp]
\centerline{\includegraphics[width=\textwidth]{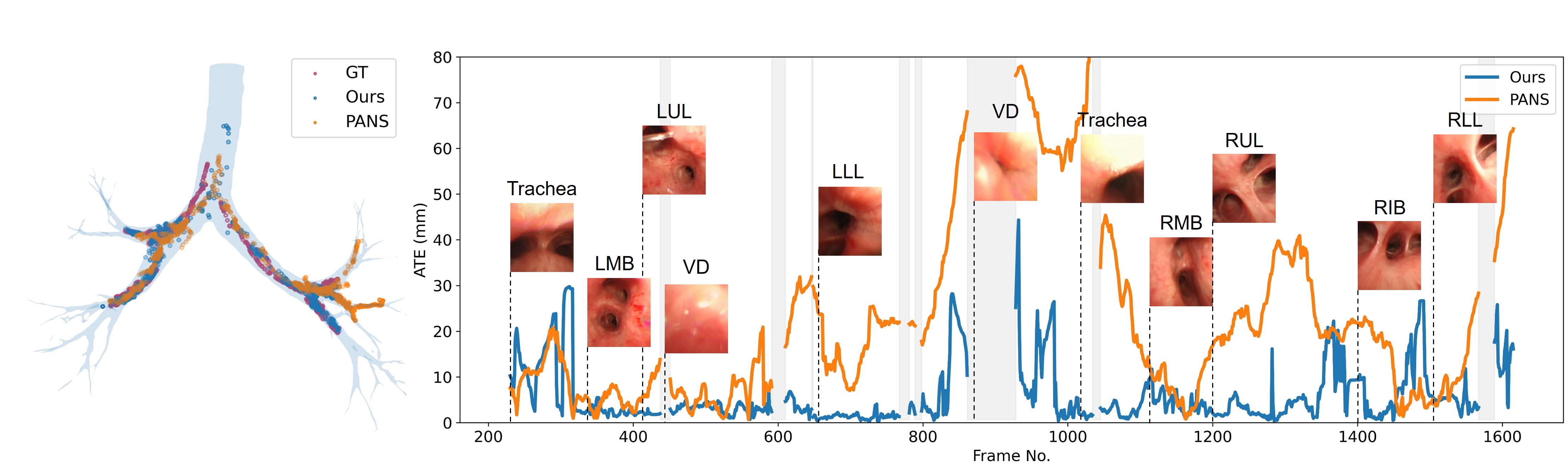}}
\caption{Localization trajectory and error for a patient case. Timesteps lacking pose labels due to severe visual degradation are shown in light gray. Example frames are provided at anatomical landmarks, including the trachea, left main bronchus (LMB), left upper lobe (LUL), left lower lobe (LLL), right main bronchus (RMB), right upper lobe (RUL), right intermediate bronchus (RIB), and right lower lobe (RLL), as well as frames affected by visual degradation (VD).}
\label{fig:comparison}
\end{figure*}

Results are presented in Table~\ref{tab:sota_comparison}. PANSv2 outperforms all compared SOTA methods, achieving an 18.1\% improvement in SR-5, which reflects a substantial gain in localization success rate. However, we observe a slight increase in ATE compared to PANS. We attribute this to occasional misidentification of anatomical landmarks, which can result in large localization errors when the bronchoscope is estimated to be in an incorrect airway branch. We also illustrate the unfiltered localization error of PANSv2 compared to PANS on a representative case. As shown in Fig.~\ref{fig:comparison}, PANSv2 rapidly recovers from visual degradation once the field of view is restored, in contrast to PANS, which fails to recover. This further demonstrates the enhanced robustness of PANSv2.

\begin{table}[t]
    \centering
    \begin{minipage}{0.49\textwidth}
        \centering
  \centering
  \caption{Comparison results across the 10-patient cases with 10,004 filtered frames.}
    \begin{tabular}{cccccc}
    \toprule
    Method & ATE (mm) $\downarrow$ & SR-5 $\uparrow$   & SR-10 $\uparrow$\\
    \midrule
    EDM   & 35.68 $\pm$ 23.23  & 3.9\% & 8.8\% & \\
    Depth-Reg & 35.18  $\pm$ 28.37  & 13.9\% & 25.7\% \\
    DD-VNB & 15.02  $\pm$ 11.68  & 22.9\% & 44.3\% \\
    PANS  & \textbf{8.68 $\pm$ 5.97} & \underline{28.4\%} & \underline{70.0}\%  \\
    Ours &  \underline{9.09 $\pm$ 11.86 } & \textbf{46.5\%} & \textbf{73.0\%} & \\
    \bottomrule
    \end{tabular}%
  \label{tab:sota_comparison}%
    \end{minipage}\hfill
    \hspace{0.1cm}%
    \begin{minipage}{0.49\textwidth}
        \centering
  \caption{Ablation results across the 10-patient cases with 15441 unfiltered frames.}
    \begin{tabular}{ccccc}
    \toprule
    Method & ATE (mm) $\downarrow$ & SR-5 $\uparrow$ & SR-10 $\uparrow$ \\
    \midrule
    w/o $L_{\text{depth}}$ & 12.87 $\pm$ 17.99  & 34.4\% & 67.2\% \\
    w/o $L_{\text{landmark}}$ & 13.70 $\pm$ 18.00  & 34.1\% & 60.9\% \\
    w/o $L_{\text{centerline}}$ & 12.76 $\pm$ 17.87  & 39.5\% & 66.0\% \\
    w/o reinit & 35.84 $\pm$ 32.13  & 17.5\% & 32.0\% \\
    Ours & \textbf{9.77 $\pm$ 12.15} & \textbf{41.8\%} & \textbf{69.7\%} \\
    \bottomrule
    \end{tabular}%
  \label{tab:ablation}%
    \end{minipage}
\end{table}

\subsection{Ablation Studies}
We evaluate the localization performance of PANS on test cases by sequentially removing the depth similarity cost, landmark alignment cost, centerline constraint, and the re-initialization module. In contrast to comparisons with SOTA methods, we present results on complete bronchoscopic videos without any filtering. As shown in Table \ref{tab:ablation}, the automatic re-initialization module significantly contributes to robust bronchoscope tracking. Fig. \ref{fig:comparison} illustrates the consistent visual degradation caused by occlusion, motion blur, and airway deformation, highlighted by the gray temporal-axis blocks. These visual artifacts pose significant challenges to airway localization methods, as the bronchoscope often continues to move despite poor visual information. By detecting tracking failures and re-initializing at identified landmarks, PANSv2 exhibits strong robustness in complex bronchoscopy scenes, quickly recovering once clear views are available. Furthermore, PANSv2 estimates pose likelihood through an integrated visual modality that combines depth similarity, landmark alignment, and centerline regularization. This approach is enhanced by foundation models with strong generalization capabilities, enabling the system to adapt to diverse anatomical variations and unseen conditions. As a result, PANSv2 achieves higher localization accuracy and stability compared to prior methods. These results underscore the effectiveness of PANSv2 framework and demonstrate its potential as a reliable tool for real-world bronchoscopic navigation.

Furthermore, we evaluate the runtime of PANSv2 and its individual components on a workstation equipped with a NVIDIA RTX 3090 GPU. The overall inference time per frame is approximately 200 ms, comprising 28 ms for depth estimation, 57 ms for landmark detection, and 111 ms for optimization. This processing speed is slower than the typical bronchoscope video capture rate (10 $\sim$ 20 Hz). The primary computational bottleneck arises from the repeated rendering given candidate bronchoscopic poses during the optimization process, which is used to assess the similarity between the rendered depth and landmark positions with the real bronchoscopic view. In future work, we aim to explore advanced optimization techniques and faster rendering engines to move the PANSv2 framework closer to real-time inference capabilities.

\section{Conclusion}
In this work, we present PANSv2, a 6-DoF bronchoscopy localization framework that leverages foundation models to enable generalizable multi-visual representation learning for improved robustness. Localization is achieved by minimizing the visual alignment cost between virtual and real bronchoscopic views using three complementary visual cues: depth similarity, landmark alignment, and centerline constraint. Combined with an automatic re-initialization module that recovers tracking during challenging scenarios, PANSv2 demonstrates state-of-the-art performance on a bronchoscopy dataset comprising 10 patient cases captured in real clinical workflow. These results highlight the effectiveness and robustness of PANSv2 in realistic and complex clinical environments, paving the way for more reliable autonomous navigation in bronchoscopy.



%

%
\bibliographystyle{splncs04}
\bibliography{references}
\end{document}